%% file: paper.tex
\documentclass[letterpaper, 10 pt, conference]{ieeeconf}  

\IEEEoverridecommandlockouts                              

\usepackage{latexsym}
\usepackage[english]{babel}
\usepackage{graphicx}
\usepackage{verbatim}

\usepackage{amsmath}
\usepackage{amsfonts}
\usepackage{relsize}

\usepackage{amsmath} 
\usepackage{amssymb}  
\usepackage{amsfonts}

\usepackage{algorithm}
\usepackage{algpseudocode}
\usepackage{epstopdf}

\usepackage{longtable}
\usepackage{textcomp}
\usepackage{url}
\usepackage{alltt}
\usepackage{caption}
\usepackage{subcaption}
\usepackage{graphicx}
\usepackage{cite}

\usepackage[table,dvipsnames,svgnames]{xcolor}

\DeclareMathOperator*{\argmin}{arg\,min}

\algnewcommand{\LineComment}[1]{\State \(\triangleright\) #1}

\newsavebox{\allttbox}

\title{CoSTAR: Instructing Collaborative Robots\\ with Behavior Trees and Vision}

\author{Chris Paxton, Andrew Hundt, Felix Jonathan, \\ Kelleher Guerin, Gregory D. Hager
  \thanks{This work was funded by NSF Grant 1227277. Authors are from the Dept. of Computer Science, Johns Hopkins University, Baltimore, MD, USA. Email: {\tt\small \{cpaxton, ahundt, fjonath1, kguerin2\}@jhu.edu} and {\tt\small hager@cs.jhu.edu}).
  }
}

\begin{document}

\maketitle




\begin{abstract}
For collaborative robots to become useful, end users who are not robotics experts must be able to instruct them to perform a variety of tasks.
With this goal in mind, we developed a system for end-user creation of robust task plans with a broad range of capabilities.
\textit{CoSTAR: the Collaborative System for Task Automation and Recognition} is our winning entry in the 2016 KUKA Innovation Award competition at the Hannover Messe trade show, which this year focused on Flexible Manufacturing.
CoSTAR is unique in how it creates natural abstractions that use perception to represent the world in a way users can both understand and utilize to author capable and robust task plans.
Our Behavior Tree-based task editor integrates high-level information from known object segmentation and pose estimation
with spatial reasoning and robot actions to create robust task plans.
We describe the cross-platform design and implementation of this system on multiple industrial robots and evaluate its suitability for a wide variety of use cases.
\end{abstract}

\section{Introduction}

While robots have not yet made inroads into homes or the world at large, collaborative robots work alongside humans in factories with increasing frequency.
These industrial robots are common in medium and large manufacturers, but are often underutilized by small manufacturers due to the high cost of retooling and reprogramming these robots to perform a wide variety of tasks.
There are two main problems with existing systems for programming these robots: clumsy user interfaces and their inability to perceive the world in ways that are meaningful to humans.
Other barriers to deployment include setup time, managing configuration details, and lack of robustness to changes in the environment.

These needs have been recognized by private enterprise.
KUKA Roboter GmbH posed the 2016 KUKA Innovation Award competition as the Flexible Manufacturing Challenge; indicating that \textit{``vision, manipulation and grasping, safe and intuitive human-robot collaboration, machine learning and cloud-based operations are considered most important''}~\cite{kuka2016award} to the future of the industry.
Our entry, \textit{CoSTAR: the Collaborative System for Task Automation and Recognition}, placed first among 6 finalists selected from 25 total applicants by a jury of robotics experts from industry and academia.
In this work, we describe CoSTAR and how it is designed to address the demand for effective collaborative robots.

\begin{figure}[bt!]
\centering
\includegraphics[width=1\columnwidth]{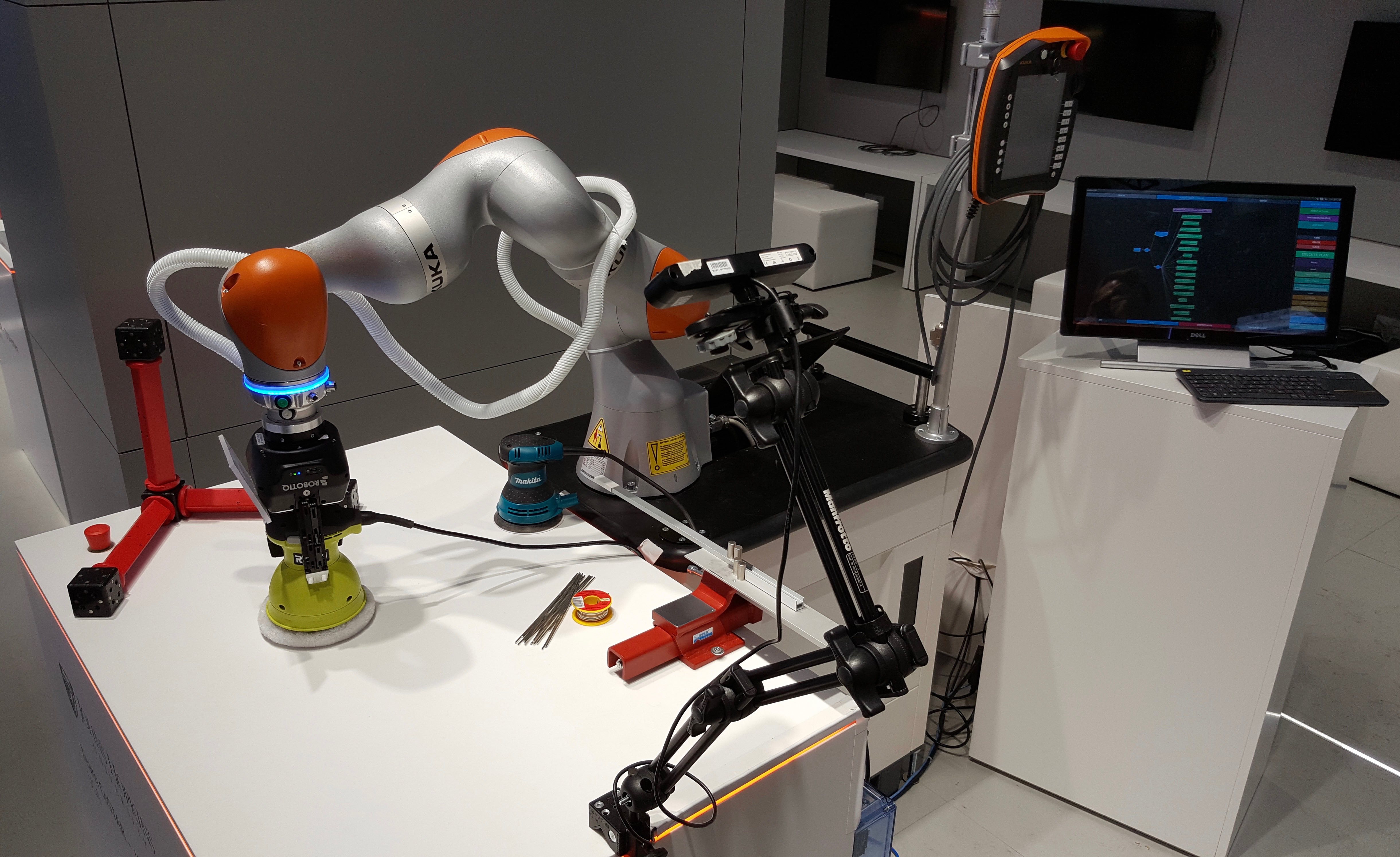}
\includegraphics[width=1\columnwidth]{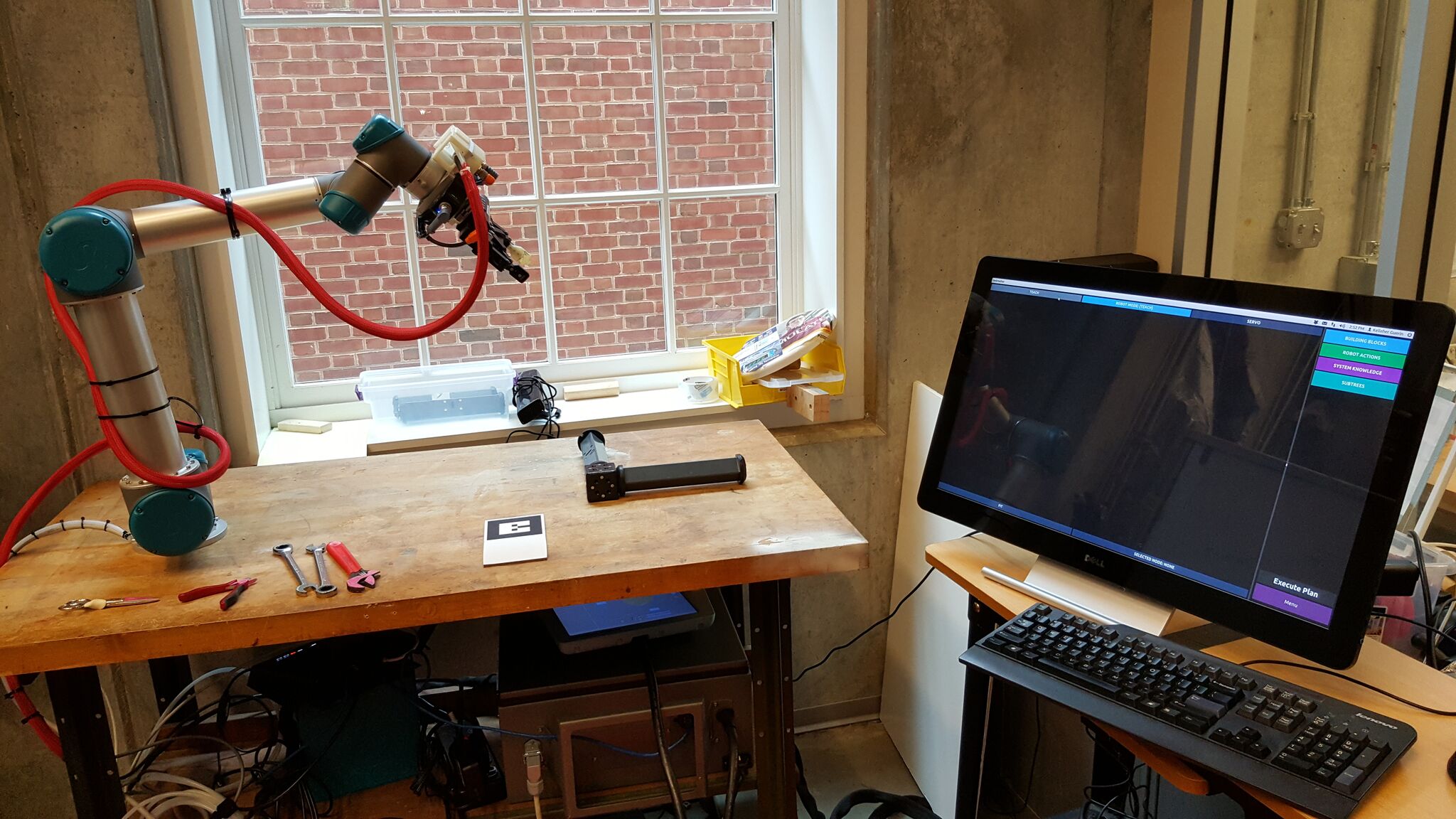}
\caption{The CoSTAR system set up to perform a wide variety of tasks.
Top: CoSTAR running on a KUKA LBR iiwa at the Hannover Messe trade show.
Bottom: CoSTAR running on a UR5 in a workshop.}
\label{fig:workspace}
\end{figure}

We have identified three characteristics key to a system for authoring robot task plans: \textit{capability}, \textit{usability}, and \textit{robustness}. 
First, a system should be capable of performing a wide variety of tasks. 
Second, end users should be able to understand the system's capabilities and efficiently create new task plans that meet their needs.
Finally, task plans should be robust to variation, and repeated executions should produce the expected result.
We designed CoSTAR to take these characteristics into consideration.

CoSTAR was originally proposed as a Behavior Tree based system to create task plans for industrial robots that utilize the tooling and human resources small manufacturers have available today~\cite{guerin2015costar}.
We extended the original system into a modular, cross-platform architecture for authoring industrial robot task plans.
We integrated perception with an abstracted world representation that allows end users to create task plans that are robust to environmental variation.
The system is shown in Fig.~\ref{fig:workspace} deployed on two different platforms.

Recent work has proposed to remedy the problems with programming robots through kinesthetic teaching methods~\cite{schou2013human}, improved graphical user interfaces (GUIs)~\cite{nguyen2013ros,mateo2014hammer,guerin2015costar}, or the use of symbols, ontologies, and natural language to enable high-level task specification~\cite{beetz2012cognition,balakirsky2012industrial,misra2014tell,fasola2014interpreting,kootbally2015towards}.
We combine a powerful GUI with grounded sensor abstractions produced by CoSTAR's distributed components. While this approach requires that users be more actively involved in constructing task plans, it results in more robust and predictable task plans~\cite{nguyen2013ros}.
CoSTAR can also rely on users' domain knowledge to solve problems without a complex ontology:
it is our philosophy that CoSTAR should empower end users to solve their own problems
rather than providing ``one size fits all'' solutions.

CoSTAR is composed of \textit{Components}, each of which is associated with input data, output data, symbols, predicates, and operations that it can perform. Input and output data are represented by ROS topics~\cite{quigley2009ros}.
Symbols represent objects and positions, while predicates describe qualities of these objects and positions.
CoSTAR includes a knowledge management component called ``Predicator,'' which 
gives end users the ability to construct a wide range of different tasks that integrate state-of-the-art object detection and pose estimation proven to work in cluttered scenes~\cite{lihierarchical}.

Our contributions are:
\begin{itemize}
\item A modular, cross-platform system designed to emphasize capability, usability, and robustness.
\item Abstract perception 
that exposes symbols and predicates that can be used for authoring task plans.
\item Evaluation of the system on a series of increasingly complex tasks.
\item Open source implementation on a KUKA LBR iiwa 14 R820 and a Universal Robots UR5, each with different grippers.
\end{itemize}

\section{Related Work}

There is broad interest in making robots into intelligent, collaborative assistants that can be taught by end users to perform complex tasks~\cite{niekum2013incremental,nguyen2013ros,mateo2014hammer}.
One approach to end user collaboration is to build tasks purely from human demonstrations. This includes work which used a BP-AR-HMM to automatically segment a task into primitive actions~\cite{niekum2013incremental}.
End-to-end deep reinforcement learning has proven effective at training individual actions~\cite{levine2016learning}.

A second approach is to allow users to provide high-level task specifications using a domain specification language such as PDDL~\cite{balakirsky2012industrial} or from natural language~\cite{fasola2014interpreting,misra2014tell}.
Balakirsky et al.~\cite{balakirsky2012industrial} used an ontology to perform a simple kitting task.
Tenorth et al. describe KnowRob, an architecture which allows robots to share knowledge, including object models and action recipes, which can be used to accomplish a variety of household chores~\cite{beetz2012cognition,tenorth2013knowrob}.
These methods provide powerful high level descriptions, but they require a large amount of built-in knowledge from an ontology which is often found to be incomplete when attempting to create a program that solves a new problem. By contrast, CoSTAR is designed to offer a suite of basic capabilities which can be recombined to solve new tasks on the fly.

A third approach is to implement an intuitive visual user interface and allow users 
to construct their own solutions to tasks. 
Mateo et al. implement Hammer for programming industrial robots, a visual programming language based on Snap running on an Android tablet~\cite{mateo2014hammer}. Nguyen et al. implemented ROS Commander as a system for building Hierarchical Finite State Machines (HFSMs) describing tasks for the PR2~\cite{nguyen2013ros}.
We expand upon the Behavior Tree visual user interface described in~\cite{guerin2015costar}.

Behavior Trees have been used to describe complex robotic manipulation tasks including a variety of object manipulation tasks ~\cite{bagnell2012integrated}, humanoid robot control~\cite{marzinotto2014towards}, and brain tumor ablation with a Raven II surgical robot~\cite{hu2015semi}.
They are comparable in power to HFSMs~\cite{colledanchise2014performance}, and are commonly used in the video game industry due to their superior scalability and modularity~\cite{isla2008halo}.
These characteristics make Behavior Trees ideal for representing collaborative robot task plans~\cite{guerin2015costar}.

We apply the algorithm described in ~\cite{lihierarchical} to object detection and pose estimation in cluttered scenes. This approach is based on ObjRecRANSAC~\cite{papazov2012rigid} with an additional step that segments objects from their environments.

\section{System Design}

\begin{figure*}[bt!]
\includegraphics[width=2\columnwidth]{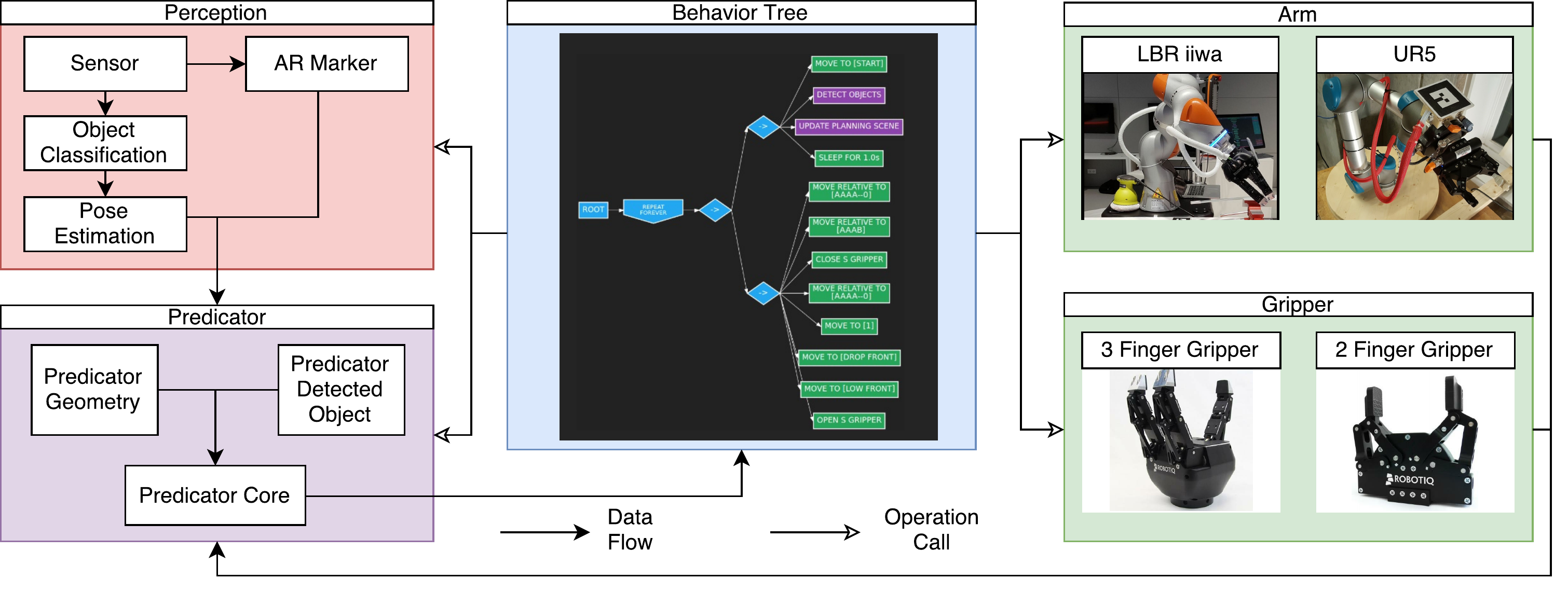}
\caption{Overview of the CoSTAR system.
Symbols and predicates are produced by components such as Perception, Arm, and Gripper, and are aggregated by Predicator.
The Behavior Tree unifies this information and uses it to encode a task, calling operations exposed by each of the individual components.
Individual components can be modified or extended to add new capabilities, but the Behavior Tree only acts on abstracted information and operations directly.}
\label{fig:costar}
\end{figure*}

We laid out three goals for CoSTAR: (1) it should have the capability of performing a wide variety of tasks, (2) it should be usable by non-experts, and (3) it should be robust to environmental change.
Incorporating perception makes the system more capable (as perception-based tasks such as sorting are now possible) and makes the system far more robust (as tasks can be performed regardless of the movement of objects and goals).
For usability, CoSTAR exposes only symbolic and qualitative information to the end user. This means that end users can formulate tasks in human terms.
In the following section, we describe the modular architecture that allows us to add new system knowledge and actions to ensure CoSTAR has the capability to perform any given task. We also describe how this connects to our usability goals via a Behavior Tree-based graphical user interface.
Our approach to end user task specification can be contrasted against approaches based on ontologies and symbolic task planning such as~\cite{balakirsky2012industrial,beetz2012cognition}.
Rather than relying on a large ontology, CoSTAR components define a relatively small set of geometric predicates such as \texttt{LeftOf}, \texttt{RightOf}, and \texttt{InFrontOf}, relying on end users to combine and use these symbols to specify tasks.

\subsection{Software Architecture}

\textbf{Components} are an extension of the system capabilities described in our previous work~\cite{guerin2015costar}, and are associated with a set of input data, output data, operations, predicates, and symbols.
Examples of CoSTAR components include the \texttt{Perception} component described in Sec.~\ref{sec:perception}, the \texttt{Gripper} component, and the \texttt{Arm} component.
Consider a CoSTAR component $C$:
\[C = <I, O, p, s, u>\]
where $p = \{p_i\}_{i=1}^N$ is the set of predicates produced by component $C$, $s = \{s_{i}\}_{i=1}^N$ is the set of symbols produced by $C$, and $u = \{u_{i}\}_{i=1}^N$ is the set of operations made possible by $C_i$. $I$ and $O$ represent continuous input and output from an individual component. Different components can be concatenations of multiple sub-components, as shown in Fig.~\ref{fig:costar}: continuous data flows between sub-components of \texttt{Perception} and \texttt{Predicator}. However, $I$ and $O$ are never explicitly exposed to the end user or the task plan.
The current world state used by the Behavior Tree is wholly defined by the set of predicates and symbols produced by all of the current components.

A key part of the modularity of the CoSTAR software design is in inheritance of components.
A particular component $C_p$ can provide a list of symbols, predicates, and operations, some of which are abstract and have not yet been implemented.
All of these must be implemented by a particular instantiation of this abstract component.
For an example of an abstract component see \texttt{Arm} in Fig.~\ref{fig:costar}: it requires implementation of basic \texttt{Teach} and \texttt{Move} operations. They must also expose an \texttt{endpoint} symbol as a coordinate frame indicating the end of the arm. Both our \texttt{LBR\_IIWA} and \texttt{UR5} components extend this \texttt{Arm} component, and implement this functionality in different ways.

\textbf{Symbols} $s$ are populated from continuous input data as a function of the raw state of the world, and represent objects, positions, and object classes. 

\textbf{Predicates} $p$ describe qualities of existing symbols and relationships between symbols. More formally, predicates are functions
\[
	p(I,s_0,\dots,s_n) \rightarrow [\texttt{TRUE}, \texttt{FALSE}]
\]
that map continuous input data and a set of symbols to a boolean value.
In effect, they discretize $I$ into meaningful subsets that can be used to create generalizable task plans.
Predicates are functions only of symbols and of the continuous input space: this prevents an explosion in predicate number and complexity. 

\textbf{Operations} $u$ are the specific actions that have an effect on the world. They can change the value of stored symbols, the state of the robot, or result in some other real-world effect. Operations typically appear as leaf nodes in the Behavior Tree; one example of an operation is \texttt{Move}.

\textbf{Predicator} is a special component which consists of multiple sub-components producing descriptive predicates, and provides operations allowing task plans to test the values of generated predicates.
Its sub-components can be activated and deactivated according to the needs of the current system.
In practice, each CoSTAR component is responsible for reporting the set of currently true predicates and the set of valid predicates and symbols to be aggregated by \texttt{Predicator}. \texttt{Predicator} then exposes operations that allow the Behavior Tree to perform queries over these predicates and symbols. 

The basic components of the CoSTAR system necessary for task plan execution are \texttt{Perception}, \texttt{Gripper}, \texttt{Arm}, and \texttt{Predicator}, as shown in Fig.~\ref{fig:costar}. 
The user interface can combine exposed operations as leaf nodes in a Behavior Tree.
Additional components (such as a \texttt{PowerTool} component for turning external tools on and off in Sec.~\ref{sec:polisher}) can be added as necessary.


For example, the \texttt{Gripper} component implements five operations, and also produces predicates describing the current state of the gripper.
All grippers can open or close. 
In addition, grippers have various modes, which determine how they are going to act when being told to open or close. 
For the more complex adaptive gripper, these are \texttt{BasicMode}, \texttt{PinchMode}, \texttt{WideMode}, and \texttt{ScissorMode}.
The parallel C-Model gripper mounted on the UR5 robot, for example, is limited to only being able to function in \texttt{PinchMode}.
Any attempt to use one of the other operations would fail.

The software was implemented in ROS~\cite{quigley2009ros}, and Orocos KDL~\cite{orocos-kdl} was used for computing robot kinematics as a part of the Arm component.
Different components expose ROS services and topics that can be run on different machines for distributed processing and execution.
The system layout, with its principal components, is given in Fig.~\ref{fig:costar}.

\subsection{User Experience}

The goal of the CoSTAR user experience is to allow users to 
teach the robot naturally, through hands-on kinesthetic teaching, and to be able to create complex task plans that can be composed quickly and visually.
The guiding principle is to allow users to teach a robot in the way they might teach a human, through a mixture of demonstration and explicit instruction.

\begin{figure}[bt!]
  \centering
  \includegraphics[width=\columnwidth]{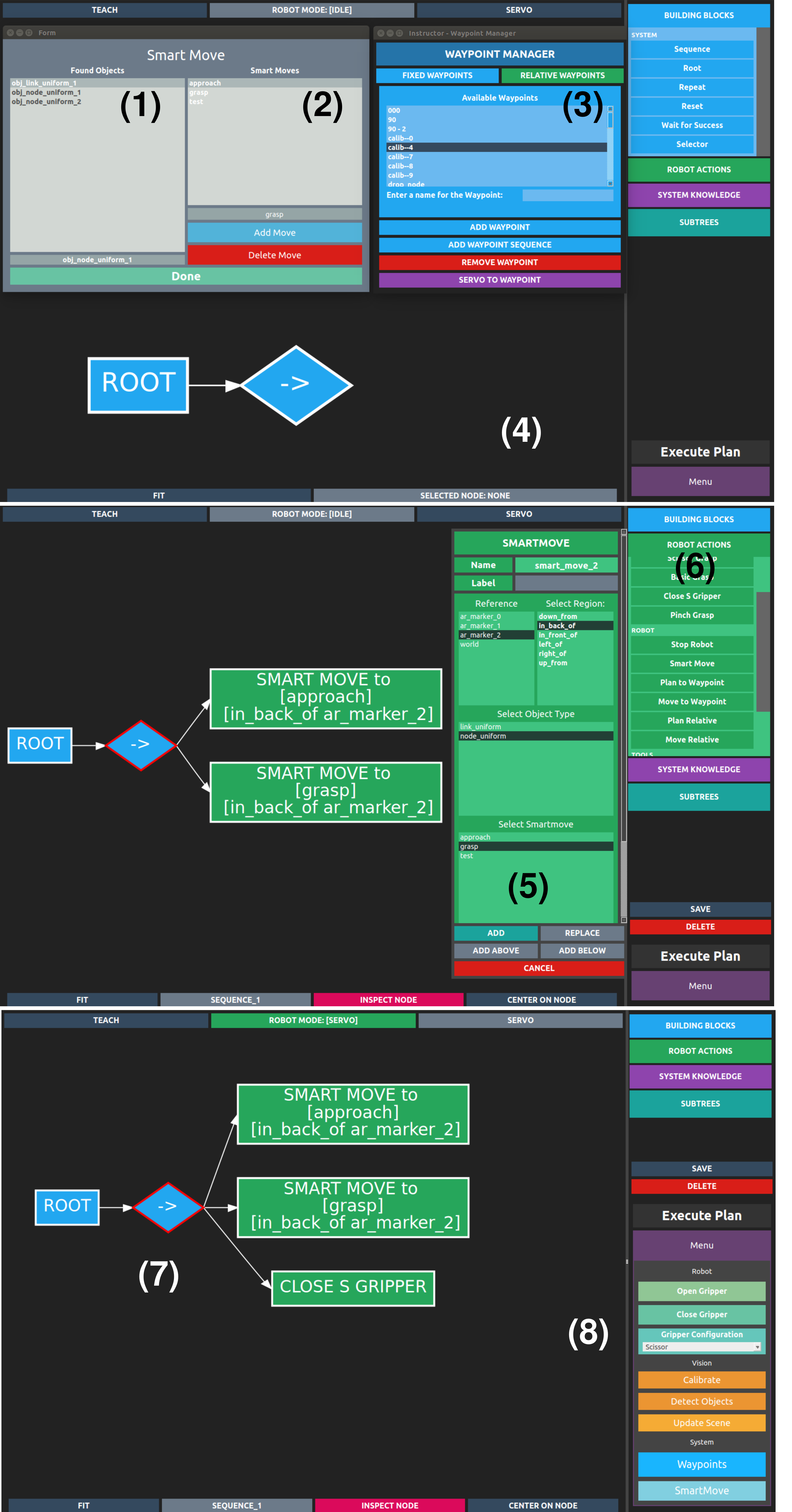}
  \caption{Behavior Tree-based user interface. 
  (1,2) Detected objects and associated SmartMoves. (3) Waypoint UI. (4) User's workspace containing the Behavior Tree. (5) SmartMove creation pane; Similar panes allow customization of other operations. (6) List of available operations. (7) Simple Behavior Tree example. (8) Expanded menu containing simple operations the user can perform during Behavior Tree development.}
  \label{fig:ui}
\end{figure}

As such we present users with two separate methods of interacting with the robot: (1) they can physically teach a robot to specify trajectories or learn skill models, and (2) they can interact with a graphical user interface that allows them to author a Behavior Tree. 
Behavior Trees are
a formalism for task construction that has been previously applied to robotics in a variety of contexts~\cite{bagnell2012integrated,guerin2015costar,hu2015semi}
that represent tasks hierarchically.
Each tree starts at a \textit{root node} which generates ticks that propagate from left to right ``down'' the tree until it reaches a leaf, which will report one of \texttt{SUCCESS}, \texttt{BUSY}, or \texttt{FAILURE}. Internal nodes control the flow of operations, and send these ticks to children according to their own internal rules and state. Operations are represented by leaf nodes in the tree. In the UI shown in Fig.~\ref{fig:ui}, internal nodes are blue, leaf nodes representing actions undertaken by the robot are green, and leaf nodes representing knowledge updates and queries are purple.

Internal nodes are key to creating complex task plans.
Examples of internal nodes are:
\begin{itemize}
\item Sequence node (\texttt{->}): tick children in order, one at a time, until each one reports \texttt{SUCCESS}. If a child fails, the sequence will fail.
\item Selection node (\texttt{?}): ticks children in order until one returns success.
\item Repeat node (\texttt{REPEAT N}): ticks children until $N$ successes or failures have been reported. 
\item Reset node (\texttt{RESET N}): returns the same value as child, but resets the child up to $N$ times.
\end{itemize}

The key advantage of behavior trees from our point of view is that they allow end users to visually create programs with the same amount of complexity and power as traditionally-written programs.
Our implementation does not include (a) user specified variables or (b) variable scope. Instead, all operations must be handled within specific nodes.
For a more in-depth examination of our implementation of the behavior tree formalism, see~\cite{guerin2015costar}.

One goal of our system is to allow end users to quickly set up the platform in a new environment. 
To facilitate this, we added a \texttt{HandEyeCalibration} component and a corresponding \textit{Calibrate} button to the user interface shown in Fig.~\ref{fig:ui}(8).
To facilitate system setup, we fix an AR marker to the end effector of the robot (visible in Fig.~\ref{fig:costar}).
We utilized dual quaternion hand eye calibration as implemented by CamOdoCal~\cite{heng2013camodocal} to compute the transform from the tip of the last joint on the robot arm's model to the marker fixed to the gripper.
After this marker transform has been computed and saved, the user can press the \textit{Calibrate} button on the CoSTAR menu to calibrate a robot to a camera as long as the arm-mounted marker is visible via this pose estimate from CamOdoCal.

\section{Perception}\label{sec:perception}

\input{perception}

\section{Evaluation}\label{sec:Evaluation}

\input{evaluation}

\section{Conclusion}

We described a modular, cross-platform system for authoring robot task plans
that is \textit{capable} of capturing a wide variety of tasks, that is \textit{usable} thanks to a powerful Behavior Tree-based user interface, and that is \textit{robust} to changes thanks to abstract perception.
We formalized individual subsystems into discrete components, each of which produces some abstracted knowledge about the world and exposes executable operations. 
Most importantly, our system grows more and more useful over time as users create task plans and develop new components. 

In the future we will perform user studies to formally assess CoSTAR's usability.
Source code for basic CoSTAR components, including the Arm and Gripper components, perception, and controlling KUKA LBR iiwa hardware is available open source as a set of ROS packages\footnote[3]{\url{https://github.com/jhu-lcsr/costar_stack}}.

\bibliographystyle{IEEEtran}
\bibliography{kuka,software,vision,costar,lfd,planners}

\end{document}

%% file: perception.tex

The key challenge of integrating perception into an interactive programming environment is to ensure the usability of the system by exposing the output and use of that perception in a way that makes sense to a human user.
In addition, perception must produce high-level knowledge that allows users to create task plans that are robust to environmental variations such as the exact positions and orientations of particular parts or manipulation goals.
All of this needs to be exposed to users in an intuitive way that allows them to build robust, comprehensible task plans.

Fig.~\ref{fig:costar} shows data flow through our perception component and to the rest of our system.
Raw RGB-D camera data is consumed by the AR marker tracker and by an object classification sub-component based on~\cite{lihierarchical}. In our case we use a Primesense Carmine.
Segmented point clouds are sent as raw input to an ObjRecRANSAC component~\cite{papazov2012rigid} which performs pose estimation, creating a set of symbols describing individual object detections. These symbols are then sent to \texttt{Predicator} which produces predicates describing those symbols in terms useful to the end user.
This behavior is exposed through a \texttt{DetectObjects} operation performed at specific known points in a task plan.


Our graphical programming system reasons over waypoints stored as 6-DOF coordinate frames 
that can come from one of three sources: (1) robot kinematics, (2) AR markers, and (3) the object detection and pose estimation pipeline.
These coordinate frames are currently produced by ObjRecRANSAC~\cite{papazov2012rigid} in the Pose Estimation step shown in Fig.~\ref{fig:costar}.

\begin{figure}[bt]
\includegraphics[width=\columnwidth]{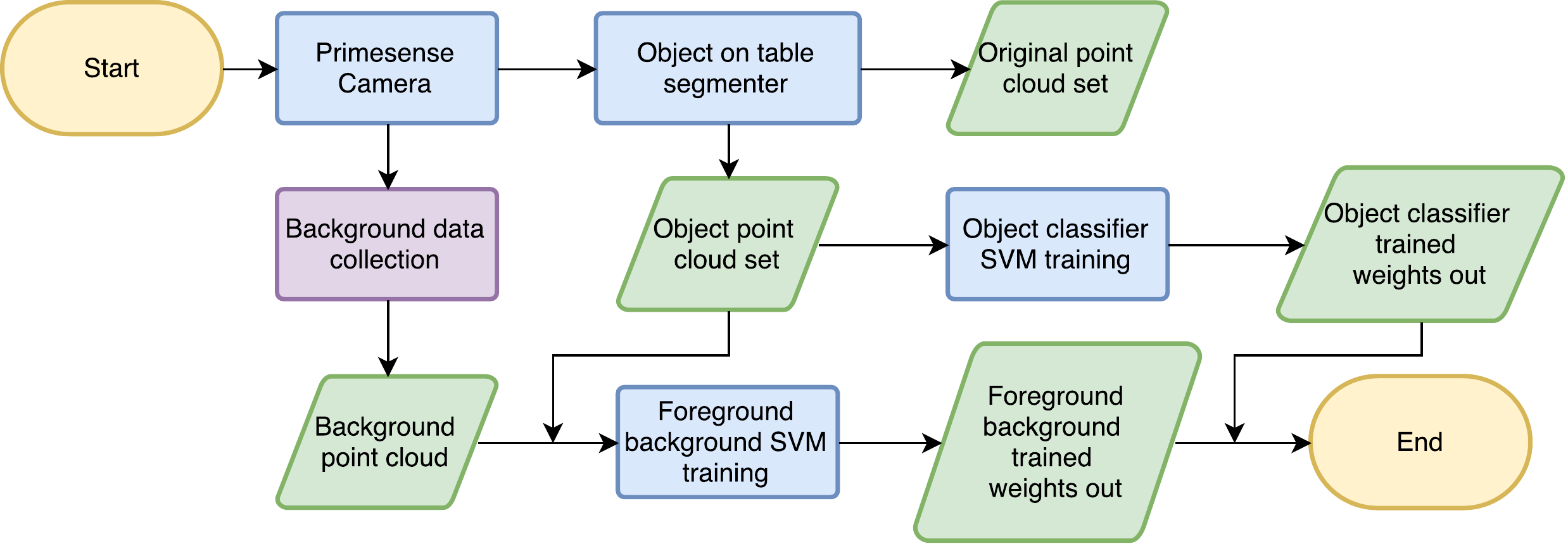}
\caption{The training algorithm for the CoSTAR perception system. This procedure is a straightforward part of system setup and needs to be performed for each new object.}
\label{fig:perception-subsystem}
\end{figure}

Most symbols used by the CoSTAR system represent a position, either expressed in 6-DOF Cartesian space relative to a world frame or a joint space coordinate.
The perception pipeline produces these symbols and associated predicates describing object class, and Predicator modules use these to assign values such as \texttt{LeftOf} and \texttt{RightOf} that describe relationships and other information pertaining to these frames.


\subsection{Object Pose Estimation Pipeline}\label{ssec:pipeline}

While the approach outlined above is effective at producing 6-DOF object pose estimates in cluttered environments, 
there are several challenges that must be handled specifically within the object pose estimation pipeline to ensure robust performance as part of a tool for authoring new task plans:

\begin{itemize}
\item It must be straightforward to adapt this system to new objects and new environments.
\item Rotational ambiguities must be resolved consistently to allow for intuitive training of grasping approaches by end users.
\item For some tasks, the identities of individual objects must be consistent across perception updates.
\item Perceptual operations must be appropriately integrated with the CoSTAR behavior tree implementation to allow construction of complex programs.
\end{itemize}

Our object detection and pose estimation pipeline reduces sensor noise with a median filter before SVM segmentation and pose estimation. 
This produces accurate estimates of coordinate frames from the frame of reference of the camera.
However, objects can contain pose ambiguities and symmetries that affect grasping angles.
For example, an axis aligned solid cube can be rotated by 90 degree increments on any of three rotation axes and remain semantically equivalent. 
We define the function \texttt{setCanonicalOrientation()} which re-orients objects in the world frame based on a canonical orientation utilizing these symmetries, a prioritized axis ordering, and the particulars of the physical object model.

To retain object identities across multiple \texttt{DetectObjects} calls, we define the function \texttt{persistenceUpdate()} uniquely number each new detection instance and enter the frame position of each object into an R*-Tree~\cite{guttman1984r,beckmann1990r} spatial data structure, ideally bulk loaded via the R-Tree packing algorithm~\cite{leutenegger1997str}.
We utilize the Boost.Geometry Spatial Index R*-Tree~\cite{boostgeometryspatialindex} implementation.
Upon each subsequent detection call a nearest neighbor lookup is performed and neighbors of the same class and within a certain threshold distance are assigned the names of previous detections.
Objects new to the scene are assigned new names.
See Alg.~\ref{alg:persistence} for the complete procedure.
This method has limitations, including velocity limits imposed by the perception update rate
and ambiguities for objects with concavities.

\begin{algorithm}[bt!]
\caption{Maintaining persistent object identities across perception updates.}
\label{alg:persistence}
\begin{algorithmic}
\State \textbf{given} R*Tree $R$, detected objects $\mathcal{O}_{detected}$,
    maximum distance $d_{max}$
\Function{persistenceUpdate}{$R,\mathcal{O}_{detected},d_{max}$}
     \State $\mathcal{O}_{persistant} = \emptyset$
     \For {$o \textbf{ in } \mathcal{O}_{detected}$}
         \State $\Call{setCanonicalOrientation}{o}$ \Comment optional line
         \State $f_{nearestNeighbor} = \Call{nearestNeighbor}{o}$
         \State $f_{within} = \Call{within}{d_{max}}$\protect\footnotemark[1]
         \State $f_{sameType} = \Call{sameType}{o}$
         \LineComment Query the R*Tree for the nearest neighbor $p$ with 
         \LineComment the same model type and within the max distance
         \State $n = R.\Call{query}{f_{nearestNeigbor} \& f_{within} \& f_{sameType}}$
        \If {$\Call{exists}{n}$}
            \State R.$\Call{remove}{n}$  \Comment enforce one match per object
        \Else
            \State $\Call{SetUniqueID}{o}$ \Comment new object
        \EndIf
     
        \State $\Call{insert}{\mathcal{O}_{persistant},o}$
     \EndFor

     \LineComment Construct R*Tree with packing algorithm~\cite{leutenegger1997str}
     \State \Return $\Call{RTree}{\mathcal{O}_{persistant}}$
\EndFunction
\end{algorithmic}
\end{algorithm}

\subsection{Perception Setup}\label{ssec:perception-training}

When deploying our system in a new environment, we need three inputs:
(1) an SVM for classifying point cloud data according to object type according to ~\cite{lihierarchical},
and
(2) 3D models for each object class identified by the SVM for use with ObjRecRANSAC~\cite{papazov2012rigid},
and (3) symmetry information for these same objects.
We outline a relatively quick data collection procedure that allows us to adapt to new environments and lighting conditions, shown in Fig.~\ref{fig:perception-subsystem}. 

The training algorithm used for the state-of-the-art object segmentation approach in~\cite{lihierarchical} requires a large amount of organized point clouds that provides multiple views of the object.
Data collection is an intuitive process that can be performed by a non-expert user requiring only an RGB-D camera and a clear workspace.
We use an AR tracking library\footnotemark[2] to specify the center of the workspace. Afterward, we do box segmentation with the center of table point cloud as the center of the box to isolate the table point cloud from its background.
We extract the points above the resulting point cloud.
This filtered point cloud is then post-processed with Euclidean clustering to separate multiple objects into several training examples that each contain a single object point cloud.
We also collect a large amount of ``negative'' data showing objects and surfaces that the system will not need to manipulate or will need to ignore, which in practice includes the robot and gripper. 
These point clouds are used in the SVM training in Fig.~\ref{fig:perception-subsystem}.


\subsection{Perceptual Operations}~\label{ssec:perception-ops}

We provide specific operations that allow users to integrate perception into a CoSTAR Behavior Tree.
The \texttt{DetectObjects} operation updates the CoSTAR system's current representation of the world from RGB-D data, 
and the \texttt{SmartMove} operation performs queries 
that select symbols representing movement goals based on 
a list of required predicates.

\footnotetext[1]{Spatial relations based on the OGC Simple Feature Specification~\cite{simplefeature}}
\footnotetext[2]{AR Track Alvar: \url{http://wiki.ros.org/ar_track_alvar}}

The \texttt{DetectObjects} operation runs the perception algorithm described in Sec.~\ref{ssec:pipeline} and updates the robot's set of known waypoints and associated predicates. After a call to \texttt{DetectObjects}, the robot will have an updated list of the position, orientation, and object type of all objects it can identify in a scene.
This is crucial for creation of \textit{robust} task plans: it creates fixed points in the program at which the robot will update its knowledge of the world.
Since users know when the \texttt{DetectObjects} operation is called, they can ensure the robot will have a non-occluded view of its workspace.



\texttt{SmartMove} is an operation that allows for intuitive end-user programming of tasks with multiple objects of the same class.
It queries \texttt{Predicator} to retrieve a list of all possible symbols matching a set of predicates.
In our implementation, users can choose a class and a geometry predicate, as shown in Fig.~\ref{fig:ui}(6).
The system then chooses a feasible motion that will take it to one of these frames according to some cost function $f$. The full algorithm is given in Alg.~\ref{alg:smartmove}.
We use this to create structure assembly and collaborative tasks in Sec.~\ref{sec:Evaluation}, where users can identify one part of the workspace for an assembled structure or human co-worker and another for the robot based on location.

\begin{algorithm}[bt!]
\caption{SmartMove operation for selecting movement goals based on predicate conditions.}
\label{alg:smartmove}
\begin{algorithmic}
\State \textbf{given} relative frame transform $T$, predicates $P$, detected objects $\mathcal{O}$, symmetries $\mathcal{S}$
\Function{SmartMove}{$T$,$P$,$\mathcal{O}$,$\mathcal{S}$}
\State $G = \emptyset$
\Comment Empty set of possible goals
\For  {$o \textbf{ in } \mathcal{O}$}
	\If {$p(o)$}
    		\For {$s_o \textbf{ in } \mathcal{S}(o)$}
    			\State \Call{Insert}{$G, o \cdot s_o \cdot T$}
    		\EndFor
	\EndIf
\EndFor
\State $g^* \leftarrow \argmin_{g \in G} f(g)$
\State \Call{Move}{$g$} \Comment Call \texttt{Arm} operation
\EndFunction
\end{algorithmic}
\end{algorithm}

%% file: evaluation.tex
As discussed in the introduction, there are three characteristics that determine the power of a framework for authoring task plans:
\begin{itemize}
\item Capability: can the system in question perform a particular task?
\item Usability: how easily can an end-user take an existing system and adapt it to a particular task?
\item Robustness: will performances of a particular task plan be the same from one trial to the next, given reasonable environmental variation?
\end{itemize}
In this section we show these characteristics of our system by demonstrating the range of tasks that can be implemented with CoSTAR on each of our robots, and discuss task plan creation and repeatability.
We published a YouTube playlist of experiments and use of our user interface.

\begin{figure*}[bt!]
\centering
\includegraphics[width=2\columnwidth]{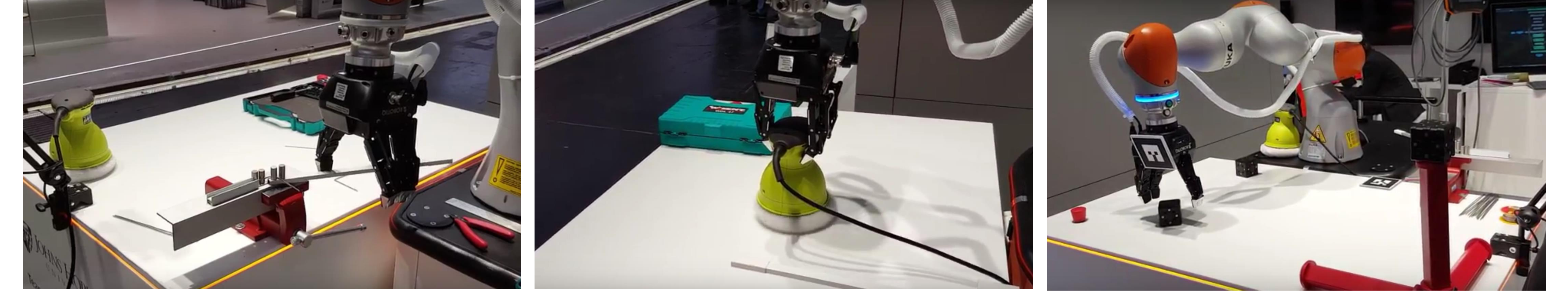}
\caption{Selected tasks implemented on the LBR iiwa using the CoSTAR system. From left to right: wire bending, polishing a surface, and collaborative structure assembly. For more videos see \url{https://www.youtube.com/playlist?list=PLCv90iHFljI3-VuVpUczGrNwvQOru3coZ}}
\label{fig:messe}
\end{figure*}

\subsection{LBR iiwa Experiments}

Our proposed system allowed us to quickly construct a wide variety of different tasks on-site at the Hannover Messe trade show.
The LBR IIWA is a very different platform from the Universal Robot UR5 we used in our other experiments: it can carry a larger gripper (in this case a Robotiq S Model 3-finger gripper) and is mounted on a mobile cart.
All tasks were constructed on site in roughly 30 minutes at the Hannover Messe trade show.
Fig.~\ref{fig:messe} shows examples of these tasks.

\subsubsection*{Wire Bending}\label{sec:wire-bending}
Our wire bending case study used a custom made wire bending jig.
This use case demonstrates roughly the same set of capabilities as the system in~\cite{guerin2015costar}: we can quickly program a set of different capabilities, but this is all ``blind'': perception is not used because the wires are too small to detect and localize with our object detection system.

\subsubsection*{Sanding and Polishing}\label{sec:polisher}
These tasks are very similar because they both use a similar tool that relies on Cartesian impedance movements. The robot must pick up the tool when it is available and move it along a known path to sand or polish some object until it is interrupted.

\begin{figure}[bt]
\includegraphics[width=\columnwidth]{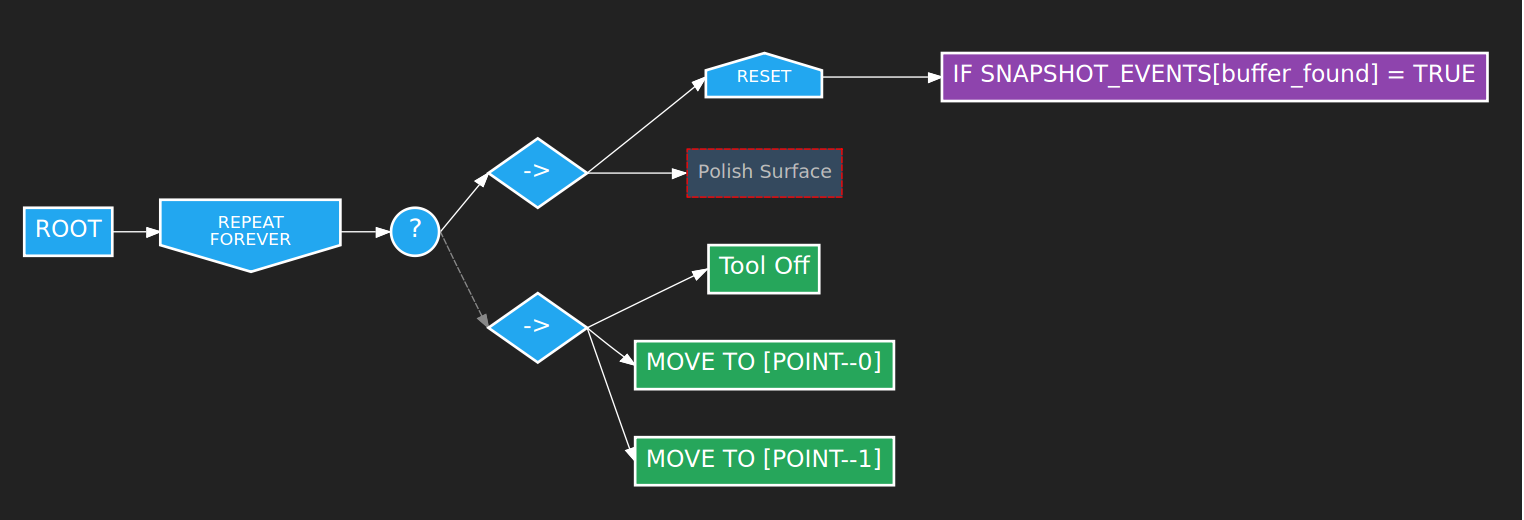}
\caption{Example of a complex task plan created for the polishing demo. The Behavior Tree includes iterator, sequence, and reset nodes.
The subtree containing the actual polishing behavior has been collapsed for readability.}
\label{fig:reset}
\end{figure}

This task demonstrates 
the integration of arbitrary external hardware into the system with an additional component that exposes \texttt{ToolOn} and \texttt{ToolOff} operations.
It also provides an example of preemption in behavior trees:
we set up a Behavior Tree that uses a selector (``\texttt{?}'') node, which means that it executes each child in order until one succeeds.
This would either tick a subtree containing a ``wait'' gesture that moved the arm up and down to indicate that the robot was ready to pick up a tool, or it would tick a subtree containing a \texttt{RESET} node guarding a polishing procedure.
The child of the \texttt{RESET} node would check to see if the tool was in position, and if this condition was false it would reset the child and return failure.
The tree is shown in Fig.~\ref{fig:reset}.

\subsubsection*{Collaborative Assembly}\label{sec:blocks}
The robot picked up nodes from the right side of a table and waited. When a human made a particular gesture, the robot would hand the node to the human user or drop it and retry. Fig.~\ref{fig:messe}(right) shows an example of this block task using the SmartMove operation described in Sec.~\ref{ssec:perception-ops}.
Here, we use the \texttt{LeftOf} and \texttt{RightOf} predicates to divide the table into an input materials and an assembly workspace. 


\subsection{UR5 Experiments}\label{ssec:ur5}


\begin{figure}[bt!]
\includegraphics[width=\columnwidth]{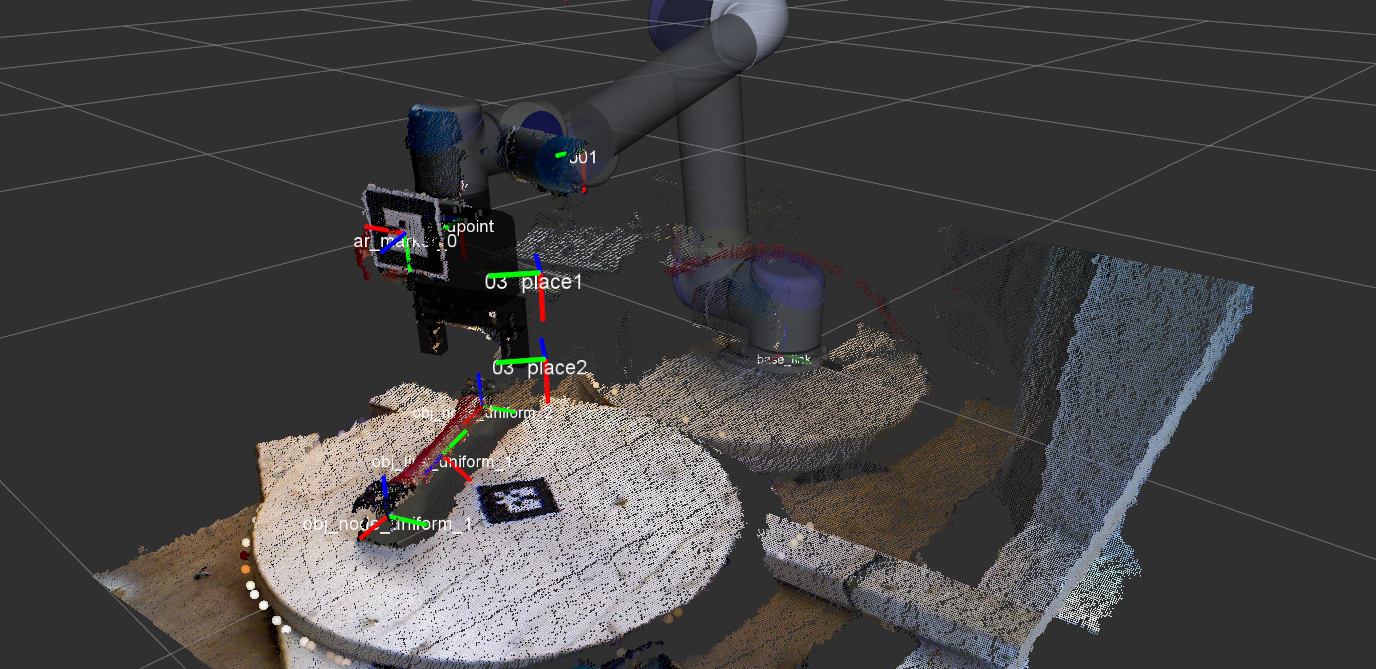}
\caption{Sensor data and object positions from UR5 experiments viewed in RVIZ.}
\label{fig:rviz}
\end{figure}

\begin{figure}[bt!]
\includegraphics[width=\columnwidth]{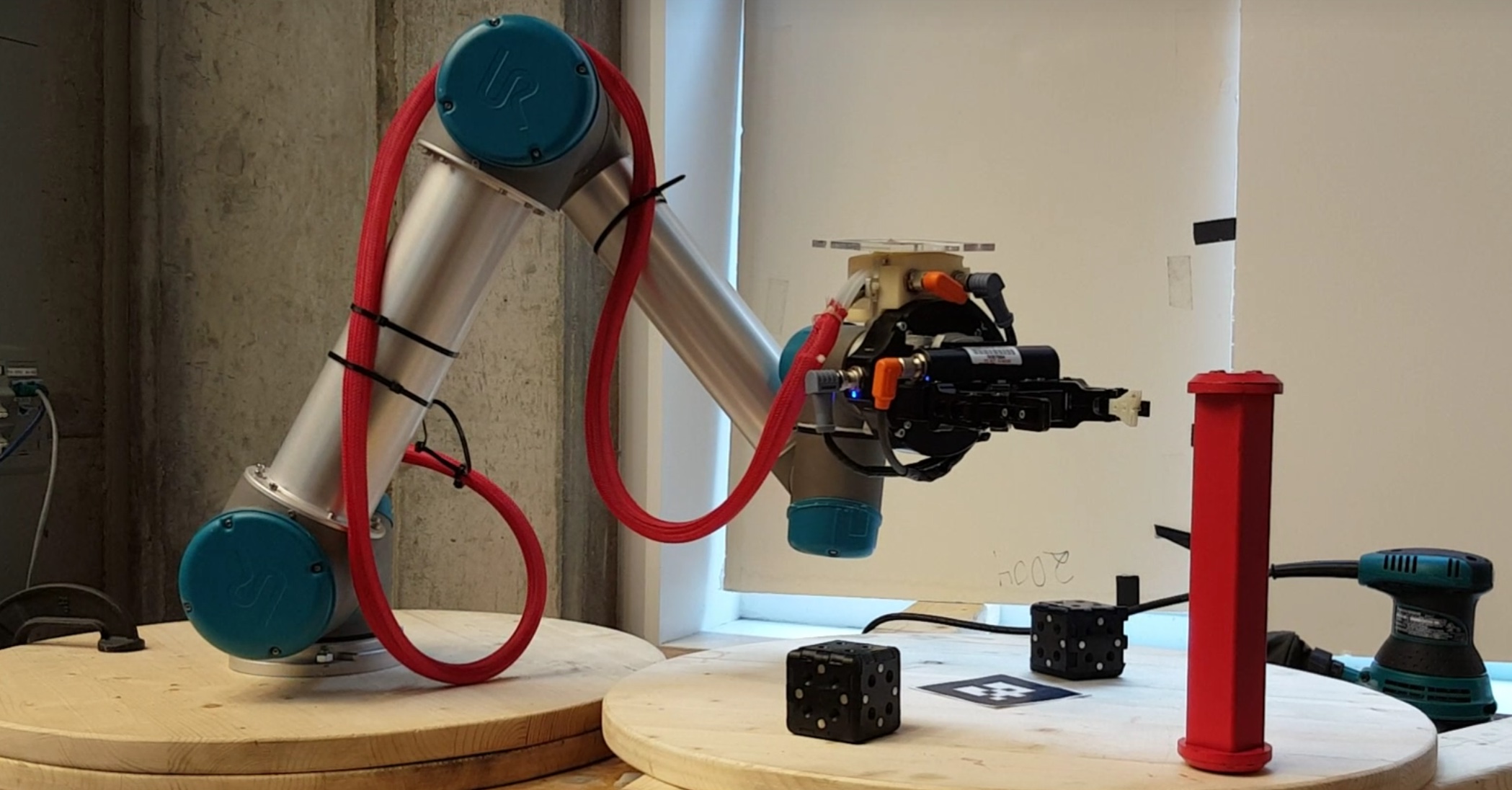}
\caption{UR5 performing assembly task.}
\label{fig:ur5}
\end{figure}


We created plans for three different tasks to demonstrate CoSTAR's ability to to construct multi-step tasks requiring precise manipulation using perception.
Figure~\ref{fig:rviz} shows the point cloud from the sensor overlaid with object pose estimates and robot joint states.
We then completed the following tasks:

\subsubsection*{1. Pick up node} First, the user programmed the UR5 to pick up any feasible node 
and lift it up.

\subsubsection*{2. Move nodes right to left} Then, the user used the existing task plan to pick up two nodes and move them from the right to the left side of the table.

\subsubsection*{3. Assemble Structure} Finally, the user modified the node-manipulation program to pick up a new node and place it at the base of a structure. After this the robot picked up a long connecting link, then finally a node to add to the task.
The experimental setup is shown in Fig.~\ref{fig:ur5}.

We performed 10 trials of the final structure assembly task, completing it successfully in 10 out of 10 trials. 
We saw no perception failures in all 10 trials, although there was one experiment a notable pose inaccuracy due to sensor noise. In this case, the robot approached the block from a bad angle but was still able to successfully grasp it and place it because the position was corrected by the closure of the gripper.
In general, we observed that barring sensor noise, the robot could complete the assembly with any set of objects for which it could find feasible grasps.